\documentclass[10pt,twocolumn,letterpaper]{article}

\usepackage{cvpr}              

\usepackage{times}
\usepackage{epsfig}
\usepackage[utf8]{inputenc}
\usepackage{url}
\newcommand{\mycomment}[1]{}
\usepackage{graphicx}
\usepackage{amsmath}
\usepackage{amssymb}
\usepackage{caption}
\usepackage{multirow, booktabs, makecell}
\usepackage[table]{xcolor}
\usepackage[accsupp]{axessibility} 
\usepackage{colortbl}
\usepackage[accsupp]{axessibility}  

\definecolor{saffron}{rgb}{0.96, 0.77, 0.19}
\definecolor{schoolbusyellow}{rgb}{1.0, 0.85, 0.0}
\definecolor{spirodiscoball}{rgb}{0.06, 0.75, 0.99}
\definecolor{paleaqua}{rgb}{0.74, 0.83, 0.9}
\definecolor{lightblue}{RGB}{221,235,247}
\definecolor{darkblue}{RGB}{31,78,120}
\definecolor{puce}{HTML}{513b41}
\definecolor{tifblue}{HTML}{c8f4f9}
\definecolor{lightbluegrey}{RGB}{230,240,255}
\definecolor{darkgrey}{RGB}{50,50,50}
\definecolor{darkgreen}{RGB}{27,64,74}
\definecolor{blue-violet}{rgb}{0.54, 0.17, 0.89}
\definecolor{byzantium}{rgb}{0.44, 0.16, 0.39}
\definecolor{darkviolet}{rgb}{0.58, 0.0, 0.83}
\usepackage[pagebackref,breaklinks,colorlinks]{hyperref}

\usepackage[capitalize]{cleveref}
\crefname{section}{Sec.}{Secs.}
\Crefname{section}{Section}{Sections}
\Crefname{table}{Table}{Tables}
\crefname{table}{Tab.}{Tabs.}

\def\cvprPaperID{19} 
\def\confName{WiCVatCVPR}
\def\confYear{2023}

\begin{document}
\title{Dense Multitask Learning to Reconfigure Comics }

\author{Deblina Bhattacharjee, Sabine Süsstrunk and Mathieu Salzmann\\
School of Computer and Communication Sciences, EPFL, Switzerland\\
{\tt\small \{deblina.bhattacharjee, sabine.susstrunk, mathieu.salzmann\}@epfl.ch}
}
\maketitle

\begin{abstract}
 In this paper, we develop a MultiTask Learning (MTL) model to achieve dense predictions for comics panels to, in turn, facilitate the transfer of comics from one publication channel to another by assisting authors in the task of reconfiguring their narratives. Our MTL method can successfully identify the semantic units as well as the embedded notion of 3D in comics panels.
This is a significantly challenging problem because comics comprise disparate artistic styles, illustrations, layouts, and object scales that depend on the author’s creative process. Typically, dense image-based prediction techniques require a large corpus of data. Finding an automated solution for dense prediction in the comics domain, therefore, becomes more difficult with the lack of ground-truth dense annotations for the comics images. To address these challenges, we develop the following solutions- we leverage a commonly-used strategy known as unsupervised image-to-image translation, which allows us to utilize a large corpus of real-world annotations; - we utilize the results of the translations to develop our multitasking approach that is based on a vision transformer backbone and a domain transferable attention module; -we study the feasibility of integrating our MTL dense-prediction method with an existing retargeting method, thereby reconfiguring comics. 
\end{abstract}

\section{Introduction}
\label{sec:intro}
\begin{figure}[ht]
\centering
{\includegraphics[width=1.0\linewidth]{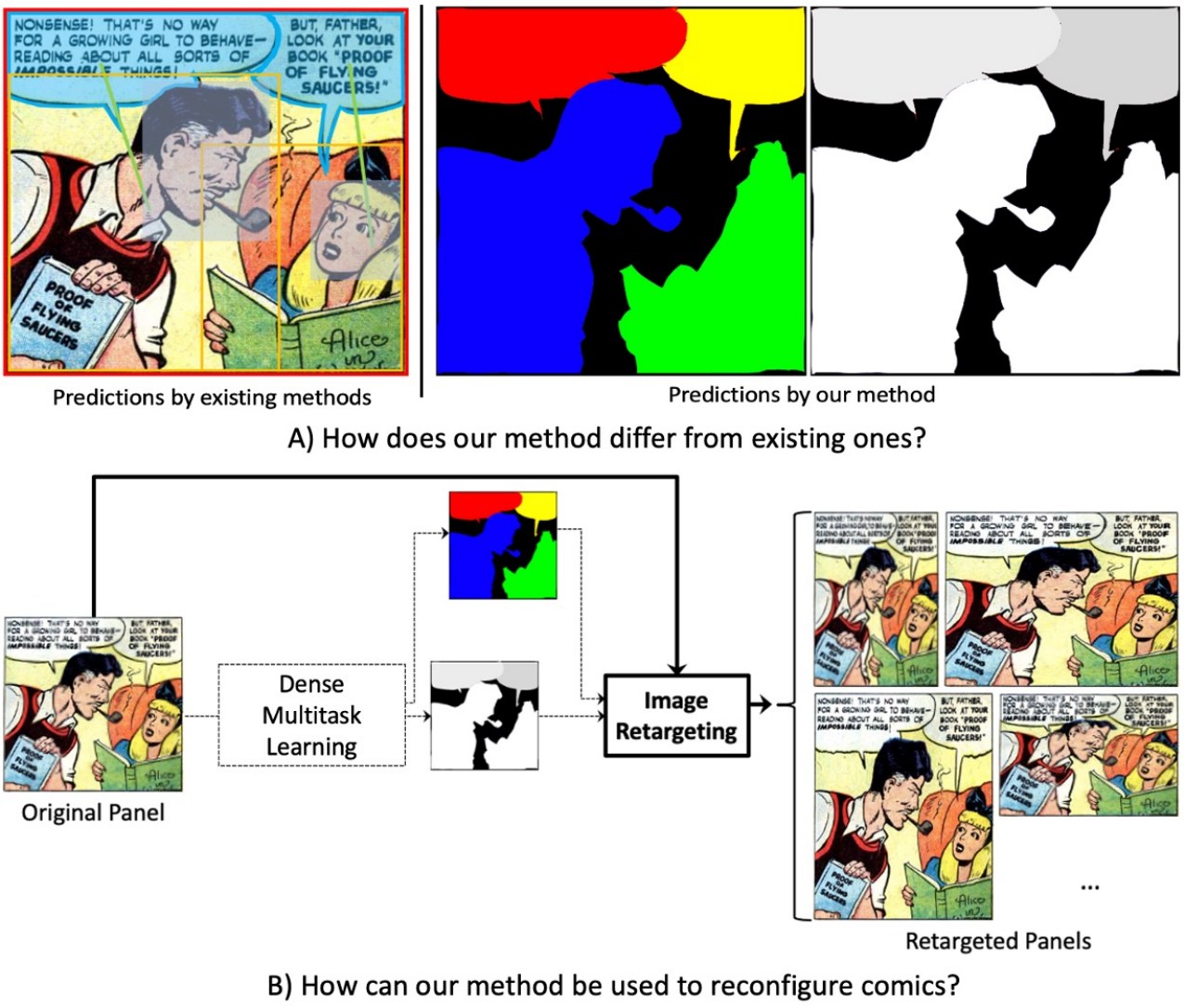}}
\caption{\textbf{Motivation for our work.} (Top) While existing methods in comics analysis can detect panels (\textcolor{red}{red} rectangle),
speech balloons (\textcolor{cyan}{cyan} mask), face (\textcolor{paleaqua}{light blue} rectangular mask), and person (\textcolor{saffron}{yellow} rectangle), we
achieve segmentation and depth predictions in a unified manner. (Bottom) Unlike the existing
methods, our dense image-based predictions can retarget comics panels, such that it can benefit comics
artists to reconfigure their work from one publication medium to another.
}\label{fig:teaser}\vspace{-10pt}
\end{figure}
Comics represent an important part of cultural heritage, preserving decades of artistic expression, stories, societal views, and lore, that predate digital media. Appreciated across age groups, the medium of comics has undergone significant evolution over the past decade. Particularly, there has been a rapid proliferation of digital comics as a consequence of reduced costs, easier transportation, and ubiquitous access. The increasing demands for comics digitization across platforms such as computers, tablets, and mobile phones, call for the automated extraction and identification of relevant elements within comics books. This process of automatically transferring the semantic or graphical units of comics from one publication medium to another is explained as \emph{reconfiguration} of comics. To achieve such reconfigurations across various publishing platforms, a viable step is investigating the relations between the comics elements by means of computational modeling.
However, this is a significantly challenging problem due to the 1) disparate styles of comics panels, 2) different text layouts, 3) changing appearances of comics characters, 4) different image scales of elements, panels, etc. 
Typically, the existing studies that investigate the various elements in comics are restricted to speech balloon segmentation~\cite{arai2011method, liu2015clump, rigaud2015speech}, text detection~\cite{rigaud2017text}, panel detection~\cite{rigaud-depth-retargeting, rigaud2013robust,  yamada2004comic}, comics character detection~\cite{chu2017manga, nguyen2017comic}, and region of interest detection~\cite{qin2017faster}. Recently,~\cite{Bhattacharjee_2022_WACV}, presented a depth estimation method on comics by exploiting the scene context. While these methods achieve promising results, they do not produce strong cues for reconfiguring the comics. Further, none of these existing techniques present a unified approach to produce multiple dense predictions of the graphical elements such as semantic segmentation and depth estimation, simultaneously. In this paper, we present a multitasking method, to segment and estimate the depth of the graphical contents within a comics panel, which are significant cues for reconfiguring comics across different media, as shown in Figure~\ref{fig:teaser}. This would help comics authors to diffuse their work across diverse publication channels, thereby benefiting the comics industry. 
Our contributions are as follows:
1) We introduce a cross-domain multitasking method to perform dense predictions by leveraging an off-the-shelf unsupervised I2I translation method and a vision transformer backbone.
2) We exploit the long-range transformer attention~\cite{swin} to achieve segmentation and depth predictions in the comics domain. To this end, we use a domain transferable attention mechanism that enforces similarity between the domains.
3) We utilize our dense MTL predictions with an existing retargeting algorithm that successfully reconfigures comics panels across different media. 
\section{Related Work}
\paragraph{Background on Comics Analysis}
The image analysis community has investigated comic book element extraction for almost 10 years, and methods vary from low-level analysis such as text recognition~\cite{arai2011method} to high-level analysis such as style recognition~\cite{chu2016manga}.
In particular,~\cite{nguyen2018digital} introduced a deep learning approach to recognize text within the panels and speech balloons by first segmenting the text areas. Differently,~\cite{chu2017manga} processed the graphical units- specifically, the comics characters by leveraging a deep learning-based detection model. Processing such graphical elements is a challenging task because of the ever-evolving appearance of the comics characters, not only across comics books but also across various panels. Nonetheless, several methods have been proposed for recognizing comic characters or faces that encompass deep neural network approaches~\cite{chu2017manga, nguyen2017comic, qin2017faster} or handcrafted feature processing techniques~\cite{khan2012color, matsui2017sketch, sun2013specific}. However, tasks like text recognition, style recognition, or character recognition do not successfully address the challenge of reconfiguring comics as they do \emph{not} produce sufficient dense pixel cues for reconfiguration. Moreover, all the existing methods- based on deep learning approaches or conventional image processing techniques- treat each element within a panel separately. In contrast, our MTL approach can achieve \emph{dense} cues that can accurately reconfigure comics images to different media automatically.

In a different vein, extracting comics panels has been extensively studied~\cite{rigaud-depth-retargeting, rigaud2013robust, yamada2004comic} to meet the increasing demands of matching the sizes of panels to that of the constantly evolving tablets, portable readers, and smartphones. While matching panel sizes can benefit the reconfiguration of comics panels, it leads to shrinking or stretching artifacts as per the device on which it is reconfigured. This means extracting panels to fit the target device sizes, does not preserve the contents or illustrations as originally intended by the authors of the comics. We mitigate this issue by identifying the key graphical elements within a panel by inferring the semantics and depth of the contents of the panel via our MTL method. 

\paragraph{Multitask Learning with Vision Transformers}
In its most conventional form, multitask learning predicts multiple outputs out of a shared encoder/representation for an input~\cite{zhang2021survey}. Prior works~\cite{taskonomy2018, zamir2020consistency, standley2019, strezoski2019taskrouting, endtoendMTL} follow this architecture to jointly learn multiple vision tasks using a CNN. Leveraging this encoder-decoder architecture, IPT~\cite{IPT} was the first transformer-based multitask network aiming to solve low-level vision tasks after fine-tuning a large pre-trained network. This was followed by~\cite{spatiotemporalMTL}, which jointly addressed the tasks of object detection and semantic segmentation. Recently, \cite{video-multitask-transformer} used a similar architecture for scene and action understanding and score prediction in videos.
Following this, Hu et.al.~\cite{hu2021unit} proposed a framework that tackles several language tasks but a single vision one. Differently, MulT~\cite{MulT} introduced a multitask transformer to handle multiple vision tasks. More complex vision transformer architectures have demonstrated that they outperform Convolutional Neural Network (CNN) based multitasking methods. However, neither do these existing methods generalize to the comics domains nor can they be trained to achieve predictions on comics imagery as they are fully-supervised networks. 

With the development of image style transfer and its connection with domain adaptation, recently~\cite{Bhattacharjee_2022_WACV} adopted style transfer and adversarial training to estimate depth on comics. In essence, the style transfer~\cite{bhattacharjee2020dunit} technique helps them to leverage models trained with large amounts of real-world ground-truth data. 
 In this vein, we apply an unsupervised I2I translation method to minimize the domain disparity between comics and the real world.
 
\subsection{Domain Adaptation via I2I Translation}
 The advent of I2I translation methods began with the invention of conditional GAN\cite{INIT26}, which have been applied to a multitude of tasks, such as scene translation~\cite{INIT13} and sketch-to-photo translation~\cite{INIT33}. While conditional GANs yield impressive results, they require paired images during training. Unfortunately, in comics$\longrightarrow$real I2I translation scenario, such paired training data is lacking and expensive to collect. To overcome this, cycleGAN~\cite{cyclegan}, with its cycle consistency loss between the source and target domains, is a possible solution for translating the comics images to real images, thereby producing consistent images. Nevertheless, neither conditional GANs, nor cycleGAN account for the multi-modality of comics$\longrightarrow$real I2I translation; in general, a single comics image can be translated to the real domain in many different, yet equally realistic ways. This is also due to the different artistic styles present in a single comics domain, which in turn, gives rise to intra-comics domain style variability. Addressing this issue of multi-modality, MUNIT~\cite{MUNIT} and DRIT~\cite{DRIT} introduced solutions by learning a disentangled representation with a domain-invariant content space and a domain-specific attribute/style space. While effective, all the above-mentioned methods perform image-level translation, without considering the object instances. As such, they tend to yield less realistic results when translating complex scenes with many objects. This is also the task addressed by INIT~\cite{INIT} and DUNIT~\cite{bhattacharjee2020dunit}. While INIT~\cite{INIT} proposed to define a style bank to translate the instances and the global image separately, DUNIT~\cite{bhattacharjee2020dunit} proposed to unify the translation of the image and its instances, thus preserving the detailed content of object instances. We, therefore, use DUNIT~\cite{bhattacharjee2020dunit} as our I2I translation model to translate the comics images to the real domain. Note that, we can also use a diffusion-based~\cite{clip} translation method to achieve comics translations. Once translated, we leverage an MTL network trained with segmentation and depth annotations from real images, to ultimately, predict the segmentation and depth of comics images.
 To enforce the domain similarity between the comics and the real domains, we utilize a transformer-based domain discriminator. We now explain our method in detail.
\begin{figure*}[ht!]
\centering
{\includegraphics[ width=1.0\linewidth ]{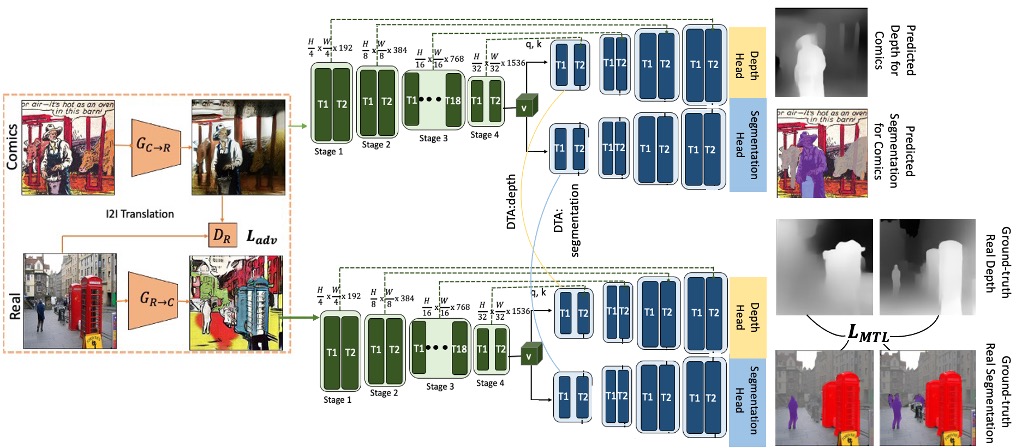}}
\caption{{\textbf{Detailed overview of our architecture for dense prediction on comics.} Our model comprises 1) an unsupervised image-to-image translation module to translate the comics images to the real domain and 2) a Swin transformer~\cite{swin} based multitasking framework that performs segmentation and depth estimation, simultaneously while bridging the domain gap using a Domain Transferable Attention (DTA). The encoder module (in \textcolor{darkgreen}{\textbf{green}}) embeds a shared representation of the input image, which is then decoded by the transformer decoders (in \textcolor{darkblue}{\textbf{blue}}) for the respective tasks. Note that the transformer decoders have the same architecture but different task heads. The MTL model is trained using a weighted loss~\cite{gradnorm} of all the tasks involved. } 
}\label{fig:detailed-model}\vspace{-10pt}
\end{figure*}

\section{Methodology}
\subsection{Problem Formulation and Overview}
We aim to learn a cross-domain multitask mapping between two visual domains $C \subset \mathbb{R}^ {H \times W \times 3}$  and  $R \subset \mathbb{R}^ {H \times W \times 3}$, where $C$ is the comics domain and $R$ is the real image domain. 
To this end, first, we employ the DUNIT model~\cite{bhattacharjee2020dunit} to translate the comics image $C$ to the real domain. This is an imperative step for achieving dense prediction on comics as we want to leverage the real-world annotations, thereby overcoming the lack of annotations in the comics domain. Moreover, simply applying an existing MTL model on these translated images does not help as comics images are highly disparate from real-world imagery. Hence, the models should be aware of the image semantic contents present in comics imagery, which is achieved by training the MTL model with the translated comics images. 

Second, we use our MTL module on the translated image. Thus, the problem can be formulated as
${TaskPred}_c = f_{\mathcal{L}_{MTL}}(R(C))$, 
where ${TaskPred}_c$ are the multitask predictions for the image in the comics domain $C$, $R(C)$ is the comics$\longrightarrow$real translated image, and $f_{\mathcal{L}_{MTL}}(R(C))$ is the MTL module trained with the multitask loss
\begin{equation}\begin{aligned}\mathcal{L}_{MTL}=w_{seg}\mathcal{L}_{CE}+w_{depth}\mathcal{L}_{rotate}\;.\end{aligned}\label{eq:mtl}\end{equation} 
on real images and applied to $R(C)$. Here, $w_{seg}$ and $w_{depth}$ are the weights of segmentation and depth tasks learned via GradNorm~\cite{gradnorm}, respectively. $\mathcal{L}_{CE}$ is cross entropy loss for the segmentation task and $\mathcal{L}_{rotate}$ is the depth loss. The detailed architecture of our method is provided in Figure~\ref{fig:detailed-model}.
We now explain the components of our network in more detail.

\subsection{Image-to-Image Translation Module}
Our method is built on the DUNIT~\cite{bhattacharjee2020dunit} backbone which embeds the input images onto a shared style space and a domain-specific content space. As such, we use the same weight-sharing strategy as DUNIT for the two style encoders $(E_c^{s},E_r^{s})$ and exploit the same loss terms. Here, $(E_c^{.},E_r^{.})$ denote the encoders in the comics and real domains, respectively. They include:
\begin{itemize}
\vspace{-5pt}
    \item A content adversarial loss ${\cal L}_{adv}^{con}(E_c^{con}, E_r^{con}, D^{con})$ relying on a content discriminator ${D^{con}}$ and the two content encoders $(E_c^{con},E_r^{con})$, whose goal is to distinguish the content features of both the comics and real domains, respectively;
    \item Domain adversarial losses ${\cal L}_{adv}^{c}(E_r^{con},E_c^{s},G_c,D^c)$ and ${\cal L}_{adv}^{r}(E_c^{con},E_c^{instcon},E_r^{s},G_r,D^r)$, one for each domain, with corresponding domain classifiers $D^c$ and $D^r$, corresponding domain generators $G_c$ and $G_r$ and instance content encoder $E_c^{instcon}$;
    \item  A cross-cycle consistency loss \\
    ${\cal L}_{1}^{cc}(G_c,G_r,E_c^{con},E_c^{instcon},E_r^{con},E_c^{s},E_r^{s})$ that exploits the disentangled content and style representations for cyclic reconstruction~\cite{cross-cc};
    \item Self-reconstruction losses\\ ${\cal L}_{rec}^c(E_c^{con},E_c^{instcon},E_c^{s},G_c)$, ${\cal L}_{rec}^r(E_r^{con},E_r^{s},G_r)$, one for each domain, ensuring that the generators can reconstruct samples from their own domain;
    \item KL losses for each domain ${\cal L}_{KL}^c(E_c^{s})$ and ${\cal L}_{KL}^r(E_r^{s})$ encouraging the distribution of the style representations to be close to a standard normal distribution;
    \item Latent regression losses ${\cal L}_{lat}^c(E_c^{con},E_c^{instcon},E_c^{s},G_c)$ and ${\cal L}_{lat}^r(E_r^{con},E_r^{s},G_r)$, one for each domain, encouraging the mappings between the latent style representation and the image to be invertible;
   \item An instance consistency loss ${\cal L}_{1}^{ic}(P_{tl}^{ci},P_{tl}^{ri},P_{br}^{ci},P_{br}^{r})$ encouraging the same object instances to be detected in the original comics image and in the corresponding image after translation, where  $P_{(.)}^{(.)}$ are the bounding box top-left and bottom-right corner pixels for detected instances in the two domains.
\end{itemize}
During training, the I2I translation module is trained separately and then we apply our MTL module. 
\vspace{-3pt}
\subsection{MTL Module}
Our MTL module follows the principle of a transformer encoder-decoder architecture~\cite{attention-is-all-you-need}. It consists of a transformer-based encoder to map the input image to a latent representation shared by the tasks, followed by transformer decoders with task-specific heads producing the predictions for the respective tasks. Figure~\ref{fig:detailed-model} shows an
overview of our MTL framework. 
For our transformer-based encoder, we use a pyramidal backbone, named the Swin Transformer~\cite{swin}
to embed the visual features into a list of hidden states that incorporates global contextual information. We then apply the transformer decoders to progressively decode and upsample the tokenized maps from the encoded image. Finally, the representation from the transformer decoder is passed to a task-specific head, such as a simple two-layer classifier (in the case of segmentation), which outputs the final predictions.  Given the simplicity of our network, it can be extended easily to more tasks. The following sections
describe the details of each component of our network.
\vspace{-10pt}
\subsubsection{Encoder Module}
For the encoder, we adopt Swin-B~\cite{swin}, which applies stacked transformers to features of gradually decreasing resolution in a pyramidal manner, hence producing hierarchical multi-scale encoded features, as shown in Figure~\ref{fig:detailed-model}. In particular, following the ResNet~\cite{resnet} structure and design rules, four stages are defined in succession: each of them contains a patch embedding step, which reduces the spatial resolution and increases the channel dimension, and a columnar sequence of transformer blocks. This approach halves the resolution and doubles the channel dimension at every intermediate stage, matching the behavior of typical fully-convolutional backbones and producing a feature pyramid (with output sizes of {1/4, 1/8, 1/16, 1/32} of the original resolution) compatible with most previous architectures for vision tasks. 

 Out of a total of $N = 24$ transformer encoders, 2 blocks are in the first, second, and fourth stages, and 18 are in the third stage. In each block, self-attention is repeated according to the number of heads used and depending on the stage of the encoding process. This is done to match the increase in the channel dimensions, where the dimensions are $M = \{6, 12, 24, 48\}$ in the first, second, third, and fourth stages, respectively. However, the high resolution in the first two stages does not allow the use of global self-attention, due to its quadratic complexity with respect to the token sequence length. To solve this issue, in all stages, the tokens, that are reshaped in a 2D representation, are divided into non-overlapping square windows of size $h = w = 7$, and the intra-window self-attention is independently computed for each of them. This means that each token attends to only the tokens in its own window, both as a query and as a key/value. A possible downside of this approach could be that the restriction to fixed local windows completely stops any type of global or long-range interaction. The adopted solution is to alternate regular window partitioning with another non-overlapping partitioning in which the windows are shifted by half their size, $\lfloor h/2 \rfloor = \lfloor w/2 \rfloor = 3$, both in the height and width dimensions. This has the effect of gradually increasing the virtual receptive field of the subsequent attention computations. 

\vspace{-10pt}
\subsubsection{Decoder Module}
Inspired by the two CNN-based decoders proposed in~\cite{SETR}, we use corresponding conceptually similar transformer-based versions. The general idea is to replace convolutional layers with windowed transformer blocks. Specifically, our decoder architecture consists of four stages, each containing a sequence of 2 transformer blocks for a total of 8. In each stage, the two sequential transformer blocks allow us to leverage inter-window connectivity by alternating regular and shifted window configurations as in the encoder.  Between consecutive stages, we use an upsampling layer to double the spatial resolution and half the channel dimension; we, therefore, adjust the number of attention heads accordingly to {48, 24, 12, 6}, in the first, second, third, and fourth stage, respectively. The spatial/channel shape of the resulting feature maps matches the outputs of the encoder stages, which are delivered to the corresponding decoder stages by skip connections. This yields an hourglass structure with mirrored encoder-decoder communication: the lower-resolution stages of the decoder are guided by the higher-level deeper encoded features and the higher-resolution stages of the decoder are guided by the lower-level shallower encoded features, allowing to gradually recover information in a coarse to fine manner and to exploit the different semantic levels where they are more relevant. 

To perform multitask prediction, we share the encoder across all tasks and use task-specific decoders with the same architecture but different parameter values. We then simply append task-specific heads to the decoder. For instance, a model jointly trained for semantic segmentation and depth prediction will have two task-specific heads: one predicting $K$ channels followed by a softmax for semantic segmentation and one predicting a single channel followed by a sigmoid for depth estimation.
\vspace{-10pt}
\begin{figure}[ht!]
\centering
{\includegraphics[width=1.0\linewidth]{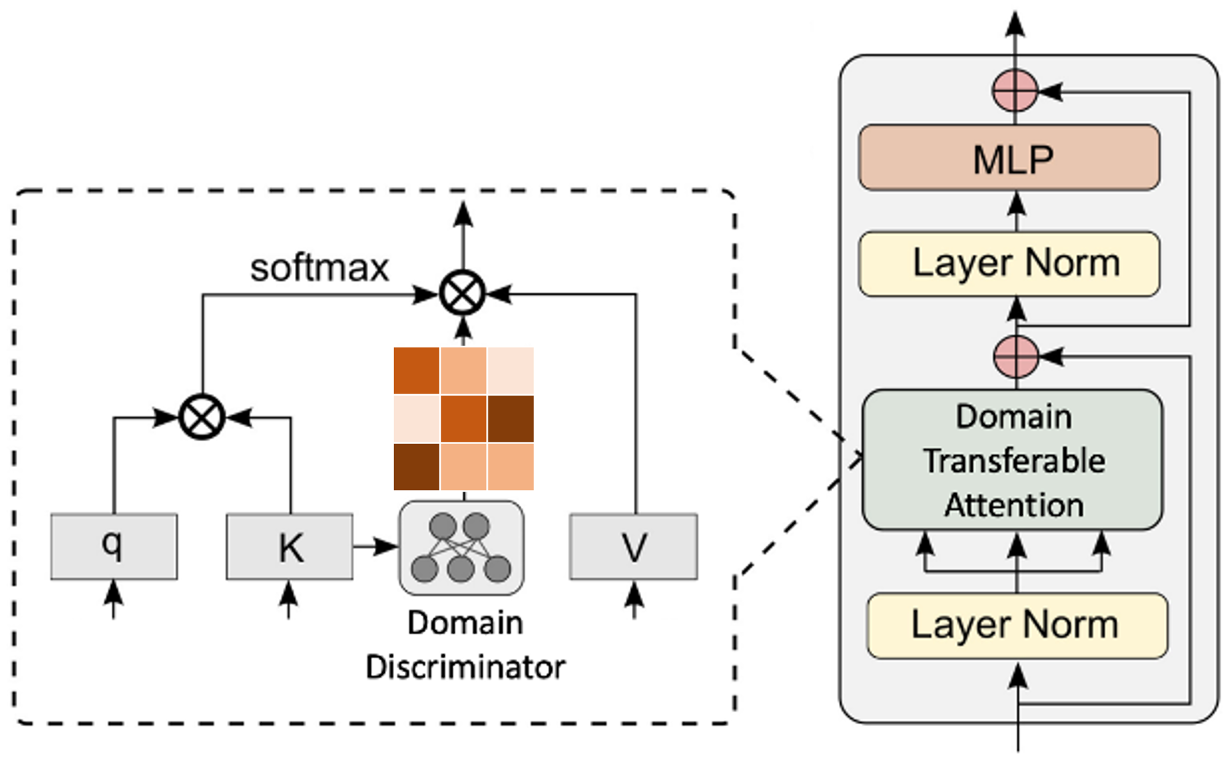}}
\caption{Overview of the \textbf{domain transferable attention} mechanism. 
}\label{fig:shared-attention-detailed}\vspace{-10pt}
\end{figure}
\vspace{-2pt}
\subsubsection{Domain Transferable Attention}
 For effective knowledge transfer between the two domains,
it is essential to focus on both transferable and discriminative features which can be leveraged from 1) the fine-grained feature tokens of Swin transformer and 2) the attention weights in the Swin transformer that convey discriminative information of the tokens between the domains.  
While the self-attention weights in Swin could be employed to discriminate between the domains, one major hurdle here is, the feature tokens do not transfer across domains. 
We, therefore, use a modified Domain Transferable Attention (DTA) mechanism~\cite{yang2021tvt} that integrates the feature tokens of the real domain into the translated comics$\longrightarrow$real domain stream. To this end, we employ a token-level domain discriminator $D_{token}$ that matches cross-domain local features by optimizing:
\begin{equation}\resizebox{\columnwidth}{!}{
$\mathcal{L}_{token}\left(x^c, x^r\right)=-\frac{1}{n} \sum_{x_1 \in C} \sum_{i=1}^n \mathcal{L}_{MTL}\left(D_{token}\left(G_f\left(x_{i }^{R\rightarrow C}\right)\right)\right)\;$.}
\end{equation}
where $n$ is the number of feature tokens, and $D_{token}\left(.\right)$ is the probability of the feature token belonging to the comics domain. 
During adversarial learning, $D_{token}$ tries to assign 1 for a feature token belonging to the comics domain and 0 for those belonging to the real domain, while $G_f$ takes the real domain feature tokens and produces fake comics tokens, denoted by $x_i^{R \longrightarrow C}$. Conceptually, a feature token that can easily deceive $D_{token}$ is more transferable across domains and should be given a higher transferability, i.e., a higher value of $D_{token}$  for a feature token in real domain implies higher transferability of that token to the comics domain. We, therefore, measure the transferability of a feature token in the real domain to that of comics by $T (.)= H\left(D_{token}\left(.\right)\right) \in[0,1]$ , where $\left.H(.)\right.$ is the standard entropy function.  This procedure allows us to leverage the tokens from the real domain that are learned via full supervision, thanks to the available real-world ground-truth annotations.  We then inject the transferable tokens from the real domain into the self-attention weights computed in the comics domain (top stream in Figure~\ref{fig:detailed-model}) as follows:
\begin{equation}\begin{aligned}\operatorname{DTA}(\mathbf{q}, \mathbf{K}, \mathbf{V})=\operatorname{softmax}\left(\frac{\mathbf{q}^T}{\sqrt{d}}\right) \odot\left[1 ; T\left(\mathbf{K}_{\text {token}}\right)\right] \mathbf{V}\;.\end{aligned}\label{eq:dta}\end{equation}

where $\mathbf{q}$ is the query of the feature token, $\mathbf{K}_{\text {token }}$ is the key of the transferable feature token from the real domain to the comics domain, $\odot$ is Hadamard product, and [;] is concatenation operation. Obviously, softmax $\left(\frac{\mathrm{qK}^T}{\sqrt{d}}\right)$ and $\left[1 ; T\left(\mathbf{K}_{\text {token }}\right)\right]$ indicate the discrimination (semantic importance) and the transferability of each token, respectively. To jointly attend to the transferability of different tasks and of different locations, we thus define DTA as:
\begin{equation}
   \begin{aligned}
\operatorname{DTA}(\mathbf{q}, \mathbf{K}, \mathbf{V}) & =\operatorname{Concat}\left(\operatorname{head}_1, \ldots, \text { head }_k\right) \mathbf{W}^o \\
\text { where head } & =\operatorname{SA}\left(\mathbf{q} \mathbf{W}_i^q, \mathbf{K W}_i^K, \mathbf{V W}_i^V\right)\;
\end{aligned} 
\label{eq:dta-head}
\end{equation}
Note that SA is the self-attention of the transformer.
This is followed by the residual and LayerNorm operation. Subsequently, an MLP and another residual operation is carried out as follows:
\begin{equation}
\begin{aligned}
\hat{z}^l & =\operatorname{DTA}\left(\operatorname{LN}\left(\mathbf{z}^{l-1}\right)\right)+\mathbf{z}^{l-1} \\ 
\mathbf{z}^l & =\operatorname{MLP}\left(\mathrm{LN}\left(\hat{\mathbf{z}}^l\right)\right)+\hat{\mathbf{z}}^l 
\end{aligned}
\label{MHA}
\end{equation}
By applying the above procedure in our MTL module, we can enforce the domains to be transferable. This, in turn, allows us to achieve dense prediction in the comics domain by leveraging supervision from real-world annotations.


Note that DTA differs from the co-attention introduced in prior works\cite{Chefer_2021_CVPR, MulT}, wherein both cases, the attention is computed based on a specific \emph{task}. By contrast, we learn a domain transferable attention between different \emph{domains}.

\vspace{-15pt}
\paragraph{Task Heads and Loss.} The feature maps from the transformer decoder modules are input to different task-specific heads to make subsequent dense predictions in the comics domain, denoted by ${TaskPred}_c$. Each class head includes a single linear layer to output a $H \times W \times 1$ map, where $H$, $W$ are the input image dimensions. We employ a weighted sum $\mathcal{L}_{MTL}$, as stated in Equation~\ref{eq:mtl}, to train the MTL network, where the losses are calculated between the ground truth and final predictions for each task in the real domain. In particular, we use cross-entropy for segmentation and rotate loss~\cite{zamir2020consistency} for depth. Note that we employ these losses to maintain consistency with the baselines~\cite{ standley2019, zamir2020consistency}.

\section{Experiments and Results}
To validate our method, we conduct experiments on the following datasets.
\subsection{Datasets}
The main dataset used for this work is the DCM dataset that comprises 772 full-page images with multiple comics panel images within. We extract 4470 single panel images from these full-page images using the panel annotations. Note that the panel annotations do not contain semantic or depth information. We thus, use these DCM panel images to translate them to real ones using~\cite{bhattacharjee2020dunit}. Following this, we apply our MTL model on the translated images. We evaluate the performance on the DCM validation set that contains dense depth annotations for 300 DCM comics images and was introduced by~\cite{Bhattacharjee_2022_WACV}. For evaluating on semantic segmentation, we use the OpenCV CVAT interface~\cite{CVAT_ai_Corporation_Computer_Vision_Annotation_2022}, leveraging the semantic labels of MS-Coco.
We further test our method for dense predictions on a novel comics test dataset of Spirou~\cite{spirou} and Tintin.
\subsection{Evaluation Metrics}
To evaluate our method, we evaluate the following two standard performance metrics, for the tasks of segmentation and depth, respectively. These metrics were reported for consistency with the baselines~\cite{unet, swin, zamir2020consistency, MulT}. \\
\textbf{Semantic segmentation} uses \emph{mIoU} as the average of the per-class Intersection over
Union (\%) between the ground-truth segmentation and predicted map.\\
\textbf{Depth} uses the Root Mean Square Error \emph{(RMSE)} computed between the depth label and the predicted depth map, where the RMSE metric is reported in meters over the evaluated set of images.

\subsection{Training Details}
We use a pretrained DUNIT module~\cite{bhattacharjee2020dunit} to first, translate the comics images to real ones. This was done for all the baselines as well. We train each baseline as per their best configurations for the tasks of segmentation and depth, mentioned in their respective works~\cite{unet, swin, zamir2020consistency, MulT}. 

We, then, train our MTL method on semantic segmentation and depth estimation. In our implementation, we train with a batch size of 8 on 2 Nvidia A100-40GB GPU, using PyTorch.
We use the weighted Adam optimizer~\cite{adam-w} with a learning rate of 5e-5 and the warm-up cosine learning rate schedule (using 2000 warm-up iterations). The optimizer updates the model parameters based on gradients from the task losses. 
\subsection{Baselines}
We compare our MTL model with the following state-of-the-art baselines.
\vspace{-15pt}
\paragraph{Baseline UNet~\cite{unet} (for single-task learning)} constitutes our CNN-based baseline. We use it as a reference for all the multitask models.
\vspace{-15pt}
\paragraph{Baseline Swin transformer~\cite{swin} (for single-task learning)} constitutes the single task transformer baseline. It is almost identical to our MTL model, except it does not include DTA, and is trained with only one dedicated task. We use it to evaluate the benefits of our multitask learning strategy
\vspace{-15pt}
\paragraph{Consistency~\cite{zamir2020consistency}} presents a general and data-driven framework for augmenting standard supervised learning with cross-task consistency. Based on a CNN backbone, it is inspired by Taskonomy~\cite{taskonomy2018} but adds a consistency constraint to learn multiple tasks jointly.
\vspace{-13pt}
\paragraph{MulT~\cite{MulT}} comprises a Swin transformer-based multitasking network with one shared encoder and multiple decoders each dedicated to a task. This baseline further identifies if tasks are interdependent, such that a shared representation can give comparable performance across multiple tasks, without explicitly adding task constraints. For MulT, we use depth as the reference task because they report the best performance with depth as a reference when jointly trained with segmentation.

Note that, we do not compare with existing works in comics analysis as none of them perform dense predictions such as segmentation or depth estimation.
All the multitask baselines were trained using their best model configurations for segmentation and depth, as in~\cite{ zamir2020consistency, MulT}, respectively. All the methods were evaluated on the translated comics image using~\cite{bhattacharjee2020dunit} for a fair comparison.

\subsection{Quantitative Results}
Table~\ref{tb:dcm-mtl} shows the comparative performance of all the evaluated baselines on the DCM validation set. Our model outperforms the multitask CNN-based baseline~\cite{zamir2020consistency} as well as the multitask Swin transformer baseline~\cite{swin} when the MTL models are jointly trained on segmentation and depth, denoted as the ’S-D’ setting. We also considerably outperform the single-task CNN baseline~\cite{unet} and the single-task Swin baseline~\cite{swin} that are trained on isolated tasks. We, therefore, report improvements in both segmentation and depth in comparison to all the baselines. Note that Tintin and Spirou are test datasets and we provide only qualitative comparisons on them. 

\begin{table}[h!]
\setlength\tabcolsep{3pt}
\centering
\scalebox{0.8}{
\begin{tabular}{lllll}
\multicolumn{3}{l}{\textbf{Quantitative results on DCM }~\cite{dcm}} 
& \multicolumn{2}{l}{~~~~~~~~\textbf{$\textit{'S-D'}$}}                                                   \\
\hline
\multicolumn{2}{c}{\multirow{-2}{*}{\textbf{Methods}}} & MTL & \cellcolor[HTML]{FFCCC9}\begin{tabular}[c]{@{}l@{}}SemSeg\\ mIoU\%$\uparrow$\end{tabular} & \cellcolor[HTML]{DAE8FC}\begin{tabular}[c]{@{}l@{}}Depth\\ RMSE$\downarrow$\end{tabular}  \\
\cmidrule(lr){1-2}\cmidrule(lr){3-3}\cmidrule(lr){4-5}
\multirow{2}{*}{CNN}& UNet~\cite{unet}  &  & 23.64
& 1.033
\\
& Cross-task Consistency~\cite{zamir2020consistency}& \checkmark&  24.72 &    0.999  
                              \\
                              \cmidrule(lr){1-2}\cmidrule(lr){3-3}\cmidrule(lr){4-5}
\multirow{3}{*}{Transformer}&1-task Swin~\cite{swin} &   & 28.95 & 0.958
                             \\

& MulT~\cite{MulT}& \checkmark   &\underline{30.00} &\underline{0.955}
                              \\
\rowcolor[HTML]{EFEFEF}
& \textbf{Our}& \checkmark  &\textbf{33.65} &\textbf{0.909} 
 \\                       \hline   
\end{tabular}}
\setlength{\abovecaptionskip}{1mm}
\caption[Quantitative results for multitasking on DCM validation set]{\textbf{Quantitative results for multitasking on DCM validation set}~\cite{dcm}. Our model outperforms both the single-task~\cite{unet, swin} and multitiask~\cite{zamir2020consistency, MulT} baselines. Bold and underlined values show the best and second-best results, respectively.}
\vspace{-10pt}
   \label{tb:dcm-mtl}%
\end{table}
\subsection{Qualitative Results}
\begin{figure}[h!]
\centering
{\includegraphics[width=1.0\linewidth]{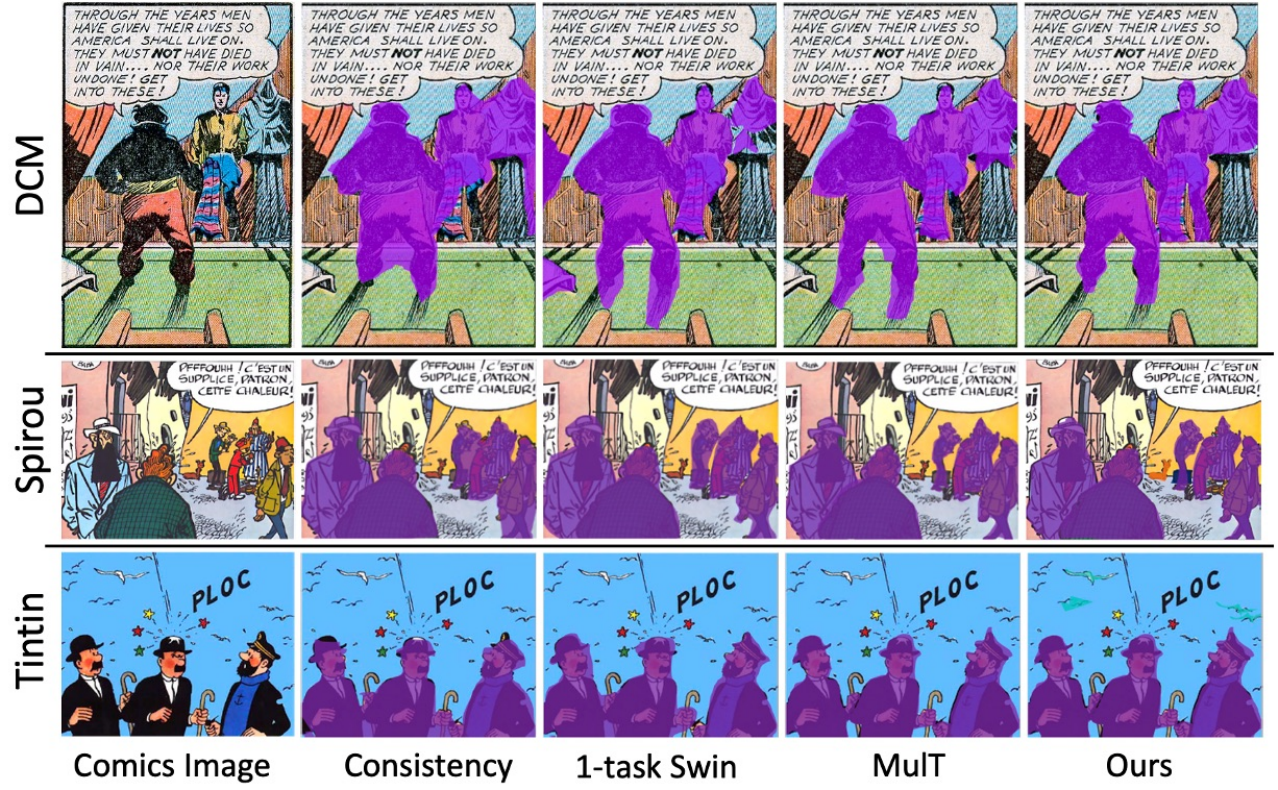}}
\caption{\textbf{Qualitative comparison of semantic segmentation} on a DCM validation~\cite{dcm} image, a Spirou test image, and a Tintin test image, respectively. We show, from left to right, the input image in the comics domain, the results using the multitask CNN-based model (Consistency)~\cite{zamir2020consistency}, the single-task Swin transformer-based segmentation model (1-task Swin)~\cite{swin}, the multitask Swin transformer-based model (MulT)~\cite{MulT}, and our model, respectively. Best viewed in color. \vspace{-10pt}
}\label{fig:qualitative-MTLresult-segmentation-dcm}
\end{figure}

\begin{figure}[h!]
\centering
{\includegraphics[width=0.9\linewidth]{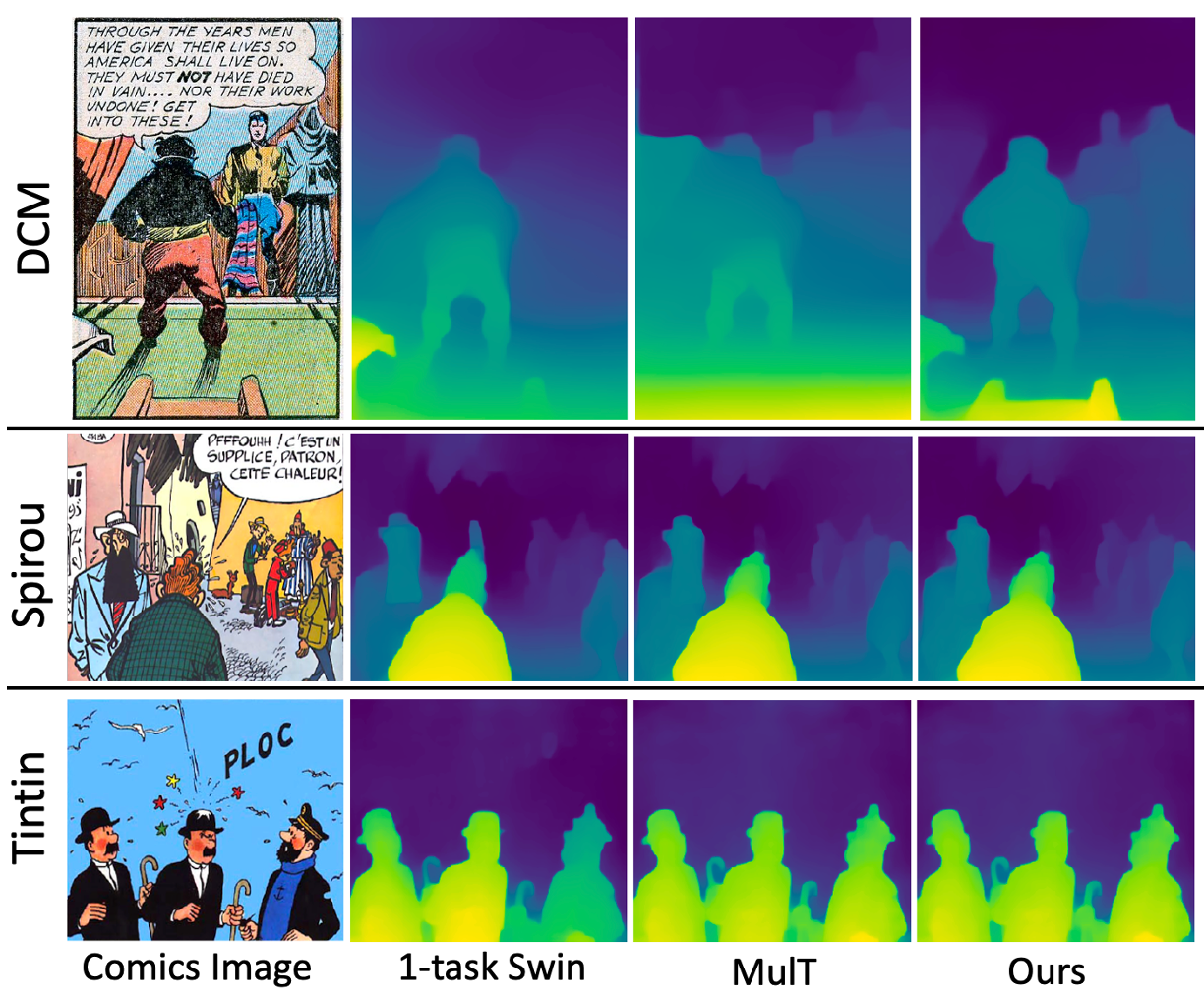}}
\caption{\textbf{Qualitative comparison of depth estimation} on a DCM validation~\cite{dcm} image, a Spirou test image, and a Tintin test image. We show, from left to right, the input image in the comics domain, the results using the single-task Swin transformer-based depth model (1-task Swin)~\cite{swin}, the multitask Swin transformer-based model (MulT)~\cite{MulT}, and our model, respectively. Best viewed in color. 
}\label{fig:qualitative-MTLresult-depth}
\vspace{-10pt}
\end{figure}

Qualitatively, we show, in Figures~\ref{fig:qualitative-MTLresult-segmentation-dcm} and ~\ref{fig:qualitative-MTLresult-depth}, the segmentation and depth results of the best-performing models from Table~\ref{tb:dcm-mtl}. The models are applied to 1) the translated DCM validation images, 2) the translated Spirou test images, and 3) the translated Tintin test images, respectively. For both segmentation and depth, we show that our method clearly outperforms the baselines, including MulT which is a transformer-based handcrafted model for MTL. Specifically, in Figure~\ref{fig:qualitative-MTLresult-segmentation-dcm}, for the Spirou test image, our method is the only approach that is able to segment the dog (shown in \textcolor{orange}{orange} segmentation mask) whereas all the baselines fail to segment it. Also, our method achieves more accurate segmentation than both the single-task as well as the multitask baselines for the 'person' category in DCM, Spirou, as well as Tintin images (shown in \textcolor{darkviolet}{purple} segmentation masks). We also show that our method achieves better foreground versus background separation in the  depth estimates in Figure~\ref{fig:qualitative-MTLresult-depth}, across all the images. Particularly, for the DCM image in Figure~\ref{fig:qualitative-MTLresult-depth}, our method accurately predicts the depth plane of the people whereas the baseline single-task Swin and the multitasking method (MulT) fail to do so.

\begin{table}[ht!]
\setlength\tabcolsep{3pt}
\centering
\scalebox{1.0}{
\begin{tabular}{lll}
\multicolumn{1}{c}{\multirow{-2}{*}{\textbf{Module}}} & \cellcolor[HTML]{FFCCC9}\begin{tabular}[c]{@{}l@{}}SemSeg\\ mIoU\%$\uparrow$\end{tabular} & \cellcolor[HTML]{DAE8FC}\begin{tabular}[c]{@{}l@{}}Depth\\ RMSE$\downarrow$\end{tabular}  \\
\cmidrule(lr){1-1}\cmidrule(lr){2-3}
I2I$+$MTL   &\underline{29.36} &\underline{0.958}
                              \\
\rowcolor[HTML]{EFEFEF}
I2I$+$MTL$+$DTA \textbf{(Our)}   &\textbf{33.65} &\textbf{0.909} 
 \\                       \hline   
\end{tabular}}
\setlength{\abovecaptionskip}{1mm}
\caption[Ablation Study on DCM validation set]{\textbf{Ablation Study on DCM validation set}~\cite{dcm}. We add the modules of our network one by one to study their effect on task performance. The DTA mechanism significantly benefits the performance by transferring the feature tokens between the real and the comics domain. Bold and underlined values show the best and second-best results, respectively.}
   \label{tb:dcm-ablation}%
   \end{table}
   \vspace{-10pt}
\subsection{Ablation Study}
In Table~\ref{tb:dcm-ablation}, we study the effect of the different components of our method. We do not isolate the I2I module as it is a necessary pre-processing step to acquire translated images. Without the translation, all the methods fail to infer comics images. Explicitly, we study the effect of the DTA mechanism on our MTL module and find that it significantly benefits dense predictions. Without the DTA mechanism, our performance is comparable to that of the handcrafted transformer-based MTL approach by MulT~\cite{MulT}.
\vspace{-10pt}
\begin{figure}[h!]
    \centering
    {\includegraphics[ width=0.65\linewidth ]{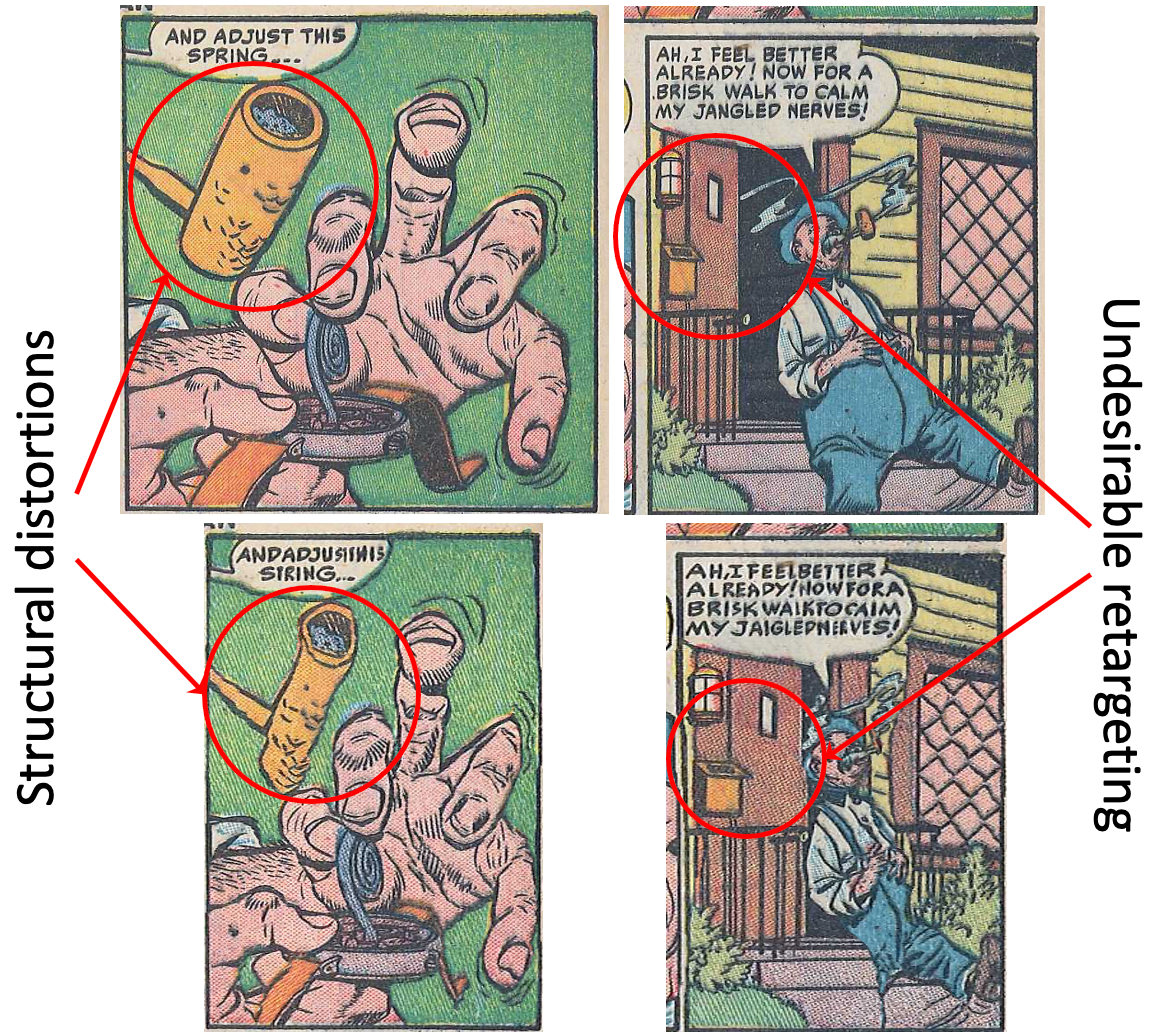}}
    \caption[Limitations of the seam-carving method on comics images]{\textbf{Limitations of the seam-carving method}~\cite{seam-carving} on comics images. }
    \label{fig:seam-carving-limitations}
    \vspace{-15pt}
\end{figure}
\subsection{Application of our MTL Method to Retargeting}
\begin{figure}[h!]
    \centering
    {\includegraphics[width=0.66\linewidth ]{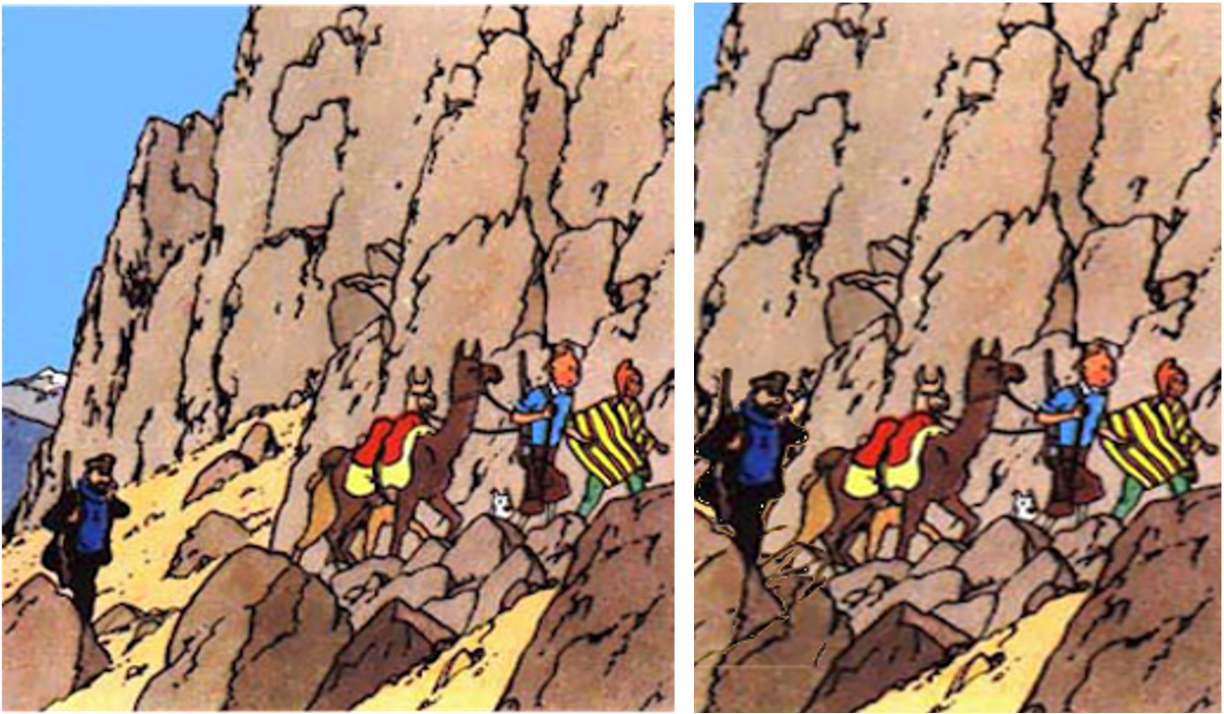}}\\    {\includegraphics[width=0.66\linewidth ]{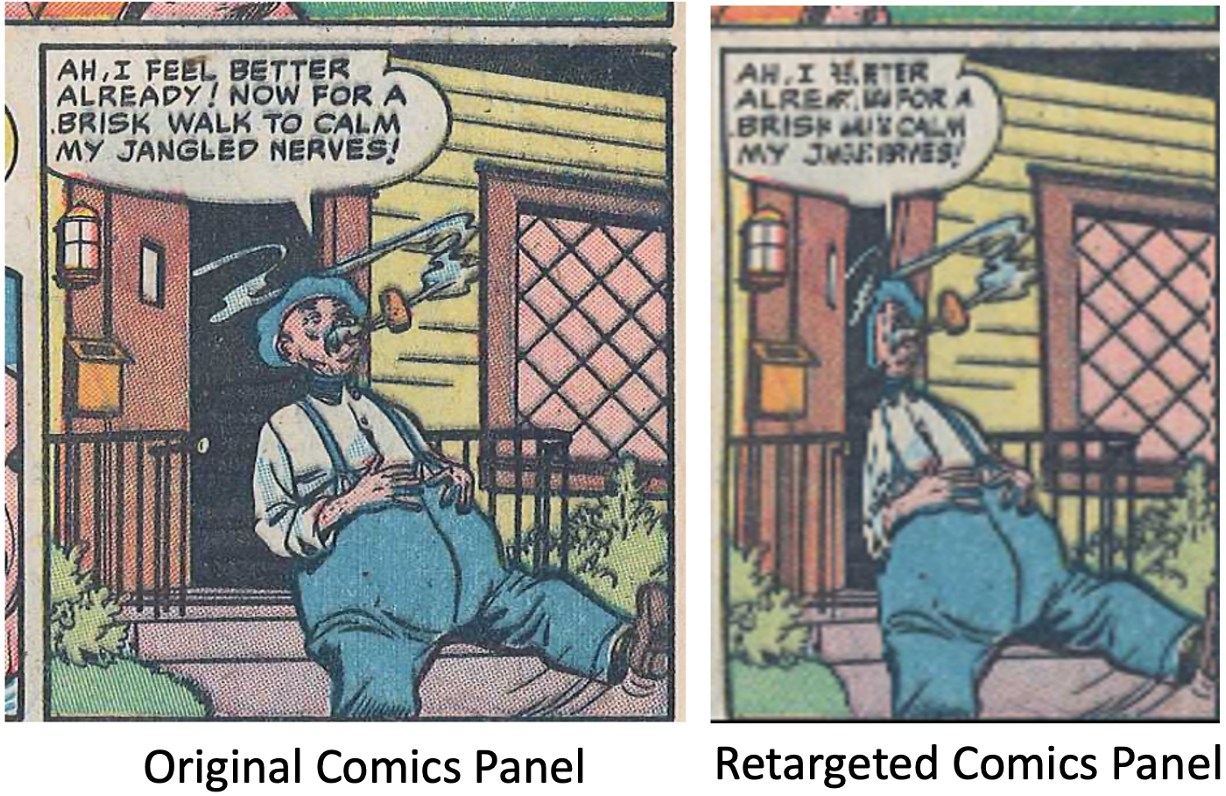}}
    \caption[Comics retargeting]{ \textbf{Applying our MTL model to retarget comics.} We integrate the dense predictions from our MTL method to guide the retargeting algorithm~\cite{transformer-retarget} to achieve results on Tintin (Top) and DCM (Bottom) test images. 
    }
    \label{fig:retarget-outline}
    \vspace{-5pt}
\end{figure}

Our MTL method can be applied to retarget comics, thereby reconfiguring them to different digital media. Particularly, existing deep retargeting methods~\cite{transformer-retarget, deep-retarget} utilize implicit semantic and depth cues to retarget semantic units accurately on the correct depth plane. However, explicitly leveraging dense prediction features to guide the learning of the retargeting method has not yet been explored. One may ask: why should one use the dense prediction cues in a deep learning framework when one can simply apply energy computations as done in the widely known seam-carving~\cite{seam-carving} method? The reason is that seam-carving, due to its non-differentiability, gives rise to structural distortions and undesirable retargeting on comics images as shown in Figure~\ref{fig:seam-carving-limitations}. Therefore, we turn to differentiable deep networks and employ additional cues from our dense task predictions to guide the retargeting methodology. In particular, we can leverage our MTL model to aid an off-the-shelf deep learning retargeting method like~\cite{transformer-retarget} to retarget comics images. The results are shown in Figure~\ref{fig:retarget-outline} where the uninformative areas of the images are removed while preserving their contents and narrative. This is one such example of retargeting comics panels to a given device size. Applying our method on a large-scale and across different media will, ultimately, help authors to reconfigure their works to diverse media in an automated manner.
\vspace{-15pt}
\section{Conclusion}
In summary, while the state-of-the-art in comics analysis remains limited to detecting panels, text, and bounding boxes for specific characters, we achieve detailed segmentation of the generic comics elements as well as infer notions of 3D from them. Benefiting from these dense predictions, this work can have different applications, such as comics scene understanding, and retargeting the comics panels. We demonstrate the applicability of our developed method by integrating it with an off-the-shelf retargeting algorithm, thereby automatically reconfiguring comics panels. This will open up possibilities to help comics authors to diffuse their work across different publication channels, thus benefiting the comics industry.
\vspace{-5pt}
\paragraph{Acknowledgement.} This work was supported in part by the Swiss National Science Foundation via the Sinergia grant CRSII5$-$180359.

 \end{document}